\documentclass[sigconf]{acmart}
\usepackage{subcaption}
\usepackage{graphicx}
\usepackage{subfloat}
\usepackage{graphicx}
\usepackage{amsmath}
\AtBeginDocument{%
  }

\copyrightyear{2023}
\acmYear{2023}
\setcopyright{acmlicensed}\acmConference[HRI '23 Companion]{Companion of the 2023 ACM/IEEE International Conference on Human-Robot Interaction}{March 13--16, 2023}{Stockholm, Sweden}
\acmBooktitle{Companion of the 2023 ACM/IEEE International Conference on Human-Robot Interaction (HRI '23 Companion), March 13--16, 2023, Stockholm, Sweden}
\acmPrice{15.00}
\acmDOI{10.1145/3568294.3580164}
\acmISBN{978-1-4503-9970-8/23/03}

\begin{document}

\title{TIP: A \textit{T}rust \textit{I}nference and \textit{P}ropagation Model in Multi-Human Multi-Robot Teams}
% Enabling Team of Teams: A Trust Inference and Propagation (TIP) Model in Multi-Human Multi-Robot Teams}


% \author{Anonymized author}
% \affiliation{%
%   \institution{Institute}
%   \city{City}
%   \state{State}
%   \country{Country}}
% \email{Email}

% \author{Anonymized author}
% \affiliation{%
%   \institution{Institute}
%   \city{City}
%   \state{State}
%   \country{Country}}
% \email{Email}

% \author{Anonymized author}
% \affiliation{%
%   \institution{Institute}
%   \city{City}
%   \state{State}
%   \country{Country}}
% \email{Email}

\author{Yaohui Guo}
\orcid{0000-0002-7064-9249}
\affiliation{%
  \institution{University of Michigan}
  \city{Ann Arbor}
  \state{Michigan}
  \country{USA}}
\email{yaohuig@Umich.edu}

\author{X. Jessie Yang}
\orcid{0000-0001-6071-0387}
\affiliation{%
  \institution{University of Michigan}
  \city{Ann Arbor}
  \state{Michigan}
  \country{USA}}
\email{xijyang@umich.edu}

\author{Cong Shi}
\orcid{0000-0003-3564-3391}
\affiliation{%
  \institution{University of Michigan}
  \city{Ann Arbor}
  \state{Michigan}
  \country{USA}}
\email{shicong@umich.edu}


\renewcommand{\shortauthors}{Guo, Yang, \& Shi}


\begin{abstract}
Trust has been identified as a central factor for effective human-robot teaming. Existing literature on trust modeling predominantly focuses on dyadic human-autonomy teams where one human agent interacts with one robot. There is little, if not no, research on trust modeling in teams consisting of multiple human agents and multiple robotic agents.
% Trust is essential for effective human-robot interaction, yet no literature studied how trust propagates through human agents in a multi-agent human-robot team. 
To fill this research gap, we present the trust inference and propagation (TIP) model for trust modeling in multi-human multi-robot teams. We assert that in a multi-human multi-robot team, there exist two types of experiences that any human agent has with any robot: direct and indirect experiences. The TIP model presents a novel mathematical framework that explicitly accounts for both types of experiences. To evaluate the model, we conducted a human-subject experiment with 15 pairs of participants ($N=30$). Each pair performed a search and detection task with two drones.  Results show that our TIP model successfully captured the underlying trust dynamics and significantly outperformed a baseline model. To the best of our knowledge, the TIP model is the first mathematical framework for computational trust modeling in multi-human multi-robot teams.
% and adopts two types of trust updates. One is the direct trust update, where a human updates trust based on his or her direct interaction with a robot; the other is the indirect trust update, where a human updates his or her trust without working directly with the robot but through his or her peer's trust. Analysis shows that a human agent's trust can stabilize under different types of repeated interactions. 

\end{abstract}

%  \input{CCSXML_code.tex}

\keywords{team of teams, multi-operator multi-autonomy (MOMA)}

\maketitle

% \vspace{-2mm}
\section{Introduction}\label{sec:intro}
Trust has been identified as one central factor for effective human-robot teaming~\cite{sheridan_humanrobot_2016, Yang2021_HFJ, yang_trust_2023}. Despite research efforts over the past thirty years, existing literature predominantly focuses on dyadic human-robot teams where one human agent interacts with one robot~\cite{NAP26355}. There is little, if not no, research on trust modeling in teams consisting of multiple human agents and multiple robots.

\begin{figure}[h]
  \begin{center}
  \vspace{0pt}
    \includegraphics[width=0.8\linewidth]{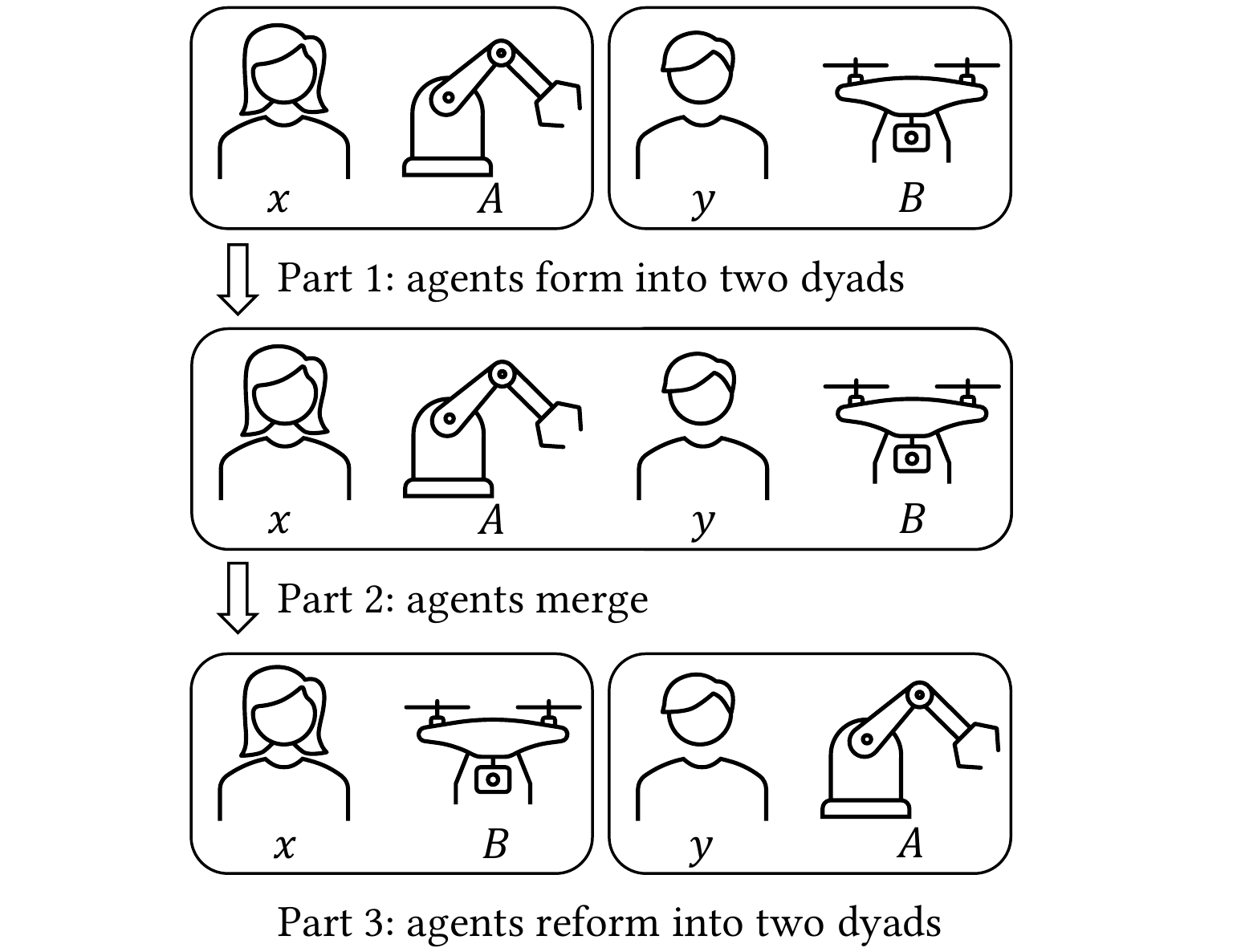}
  \end{center}
  \vspace{-8pt}
  \caption{Four agents can form sub-teams. In Part 1, human $x$ and robot $A$ form a dyad, and human $y$ and robot $B$ form a dyad. In part 2, two dyads merge. In part 3, human $x$ and robot $B$ form a dyad, and human $y$ and robot $A$ form a dyad.}
  \label{fig:scenario}
\end{figure}

Consider a scenario where two human agents, $x$ and $y$, and two robots, $A$ and $B$, are to perform a task. The four agents are allowed to form sub-teams to enhance task performance (e.g., maximizing throughput and/or minimizing task completion time). For instance, they could initially form two dyadic human-robot teams to complete the first part of the task, merge to complete the second part, and split again with a different configuration to complete the third part of the task, and so on (see Fig \ref{fig:scenario}). 

  \vspace{-2pt}
  
\begin{figure}[h]
  \centering
  \includegraphics[width=1\linewidth]{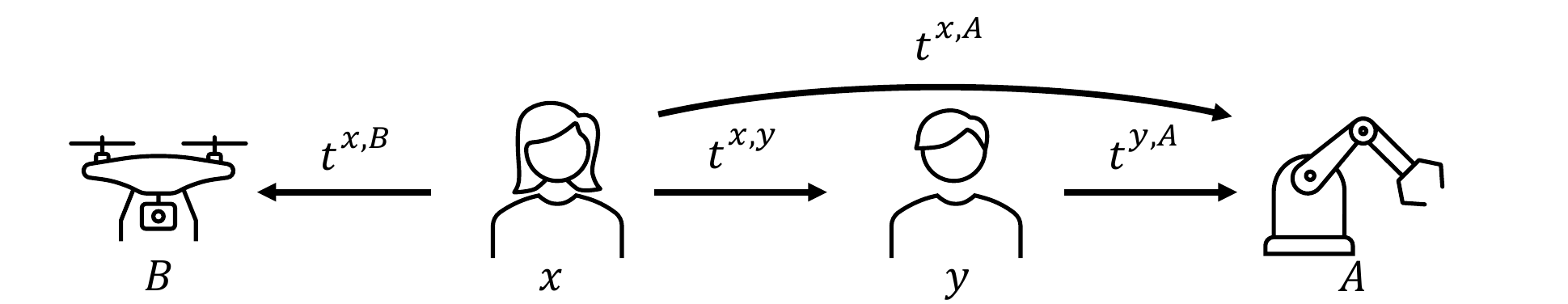}
  \vspace{-12pt}
  \caption{An arrow points from a trustor to a trustee, representing the trust $t^{\text{trustor}, \text{trustee}}$. Human $x$ updates her trust in robot $B$ via direct experience. 
  % $x$ uses learns human teammate $y$'s trust in $A$, $t^{y,A}_k$ and update her trust in $A$, $t^{x,A}_k$. 
  Even though $x$ does not have direct interaction with $A$, $x$ could still update his or her trust toward $A$ through a third party, $y$. }
  \label{fig:trust-transfer}
\end{figure}

In this scenario, we assert that there exist two types of experiences that a human agent has with a robot: \emph{direct and indirect experiences}. Direct experience, by its name, means that a human agent's interaction with a robot is by him-/her-self; indirect experience means that a human agent's interaction with a robot is mediated by another party. Fig.~\ref{fig:trust-transfer} illustrates Part \emph{3} of the task shown in Fig. \ref{fig:scenario}. Human $x$ works directly with robot $B$ (i.e., direct experience). Even though there is no direct interaction between $x$ and $A$ in part 3, we postulate that $x$ could still update his or her trust in $A$ by learning $y$'s experience with $A$, i.e., $y$'s direct experience with $A$ becomes $x$'s indirect experience. $y$'s trust in $A$ \emph{propagates} to $x$. 

Under the direct and indirect experience framework, prior work on trust modeling in dyadic human-robot teams can be regarded as examining how \emph{direct} experience influences a person's trust in a robot. In multi-human multi-robot teams, we postulate that \textit{both direct and indirect experiences drive a human agent's trust in a robot}.

In this study, we develop the \textbf{T}rust \textbf{I}nference and \textbf{P}ropagation (TIP) model for multi-human multi-robot teams, explicitly accounting for the direct and indirect experiences a human agent may have with a robot. We examine trust dynamics under the TIP framework and prove theoretically that trust converges after repeated (direct and indirect) interactions. To evaluate the proposed TIP model, we conducted a human-subject experiment with 15 pairs of participants ($N=30$). Each pair worked with two drones to perform a threat detection task for 15 sessions. We compared the TIP model (i.e., accounts for both the direct and indirect experiences) and a direct-experience-only model (i.e., only accounts for the direct experience a human agent has with a robot). Results show that the TIP model successfully captured people's trust dynamics with a significantly smaller root-mean-square error (RMSE) compared to the direct-experience-only model. To the best of our knowledge, the proposed TIP model is the first mathematical framework for computational trust modeling in multi-human multi-robot teams.

\section{Related Work}\label{sec:relatedWork}
Several computational trust models in dyadic human-robot teams exist \cite{chen2018planning,  Xu2015optimo,  Guo2020_IJSR, pippin2014trust}. Notably, Xu and Dudek~\cite{Xu2015optimo} proposed the online probabilistic trust inference model (OPTIMo) utilizing Bayesian networks to estimate human trust from automation's performance and human behavior. Hu et al.~\cite{hu2016real} proposed to classify trust or distrust based on electroencephalography (EEG) and galvanic skin response (GSR) signals. Soh et al.~\cite{soh2020multi} modeled trust as a latent dynamic function for predicting human trust in robots across different tasks. Guo and Yang~\cite{Guo2020_IJSR} and Bhat et al. \cite{Bhat_RAL_2022} proposed to model trust as a Beta random variable and predicted trust value in a Bayesian framework. For a detailed review, refer to \cite{kok2020trust}.

Even though the amount of research on trust modeling in multi-human multi-robot teams is extremely limited, some inspiration can be drawn from research on reputation/trust management. Central to the reputation management system is a propagation mechanism that allows a buyer to obtain the reputation/trustworthiness values of a seller, especially when the buyer had no prior transactions with the seller~\cite{hendrikx2015reputation}.
% , where trust can propagate through the network. 
Examples include the Beta reputation system~\cite{ josang2002beta} and the FIRE trust management model~\cite{huynh2004fire}. Variants of such propagation mechanisms can be found in several multi-agent systems, such as e-commerce~\cite{cen2019trust} and social networks~\cite{fan2019graph}.

% \vspace{-2mm}
\section{Mathematical Model}\label{sec:model}
We present the detailed TIP model. Our objective is to develop a fully computational trust propagation model that works in general human-robot interaction settings. 

\vspace{-2mm}
\subsection{Assumptions}\label{subsec:assumptions}
We make two major assumptions. First, we assume that each human agent communicates trust as a single-dimensional value \cite{Guo2020_IJSR, guo2021reverse}. Second, we assume that human agents are cooperative, i.e., they report their trust truthfully to their human teammates. 

\vspace{-2mm}
\subsection{Proposed Model}\label{sec:model}

\noindent \textbf{Trust as a Beta random variable.} We define a trust value as a real number in $[0,1]$, where 0 stands for ``[do] not trust at all'' and 1 stands for ``trust completely''. We take a probabilistic view to model trust as in ~\cite{Guo2020_IJSR}. At time $k$, the trust $t_{k}^{a,b}$ that a human agent $a$ has towards another agent $b$ follows a Beta distribution, i.e., $t_{k}^{a,b} \sim \operatorname{Beta}\left( \alpha _{k}^{a,b} ,\beta _{k}^{a,b}\right)$,
% \begin{equation} \label{eq:trust_def} 
% % t_{b}^{a} (k)\sim \operatorname{Beta}\left( \alpha _{b}^{a} (k),\beta _{b}^{a} (k)\right)
% t_{k}^{a,b} \sim \operatorname{Beta}\left( \alpha _{k}^{a,b} ,\beta _{k}^{a,b}\right)
% \text{,}
% \end{equation}
where $\alpha _{k}^{a,b}$ and $\beta _{k}^{a,b}$ are the positive and negative experiences $a$ had about $b$ up to time $k$, respectively, $k=0,1,2,\dots$. When $k=0$, $\alpha _{0}^{a,b}$ and $\beta _{0}^{a,b}$ are the prior experiences that $a$ had before any interaction with $b$. The expected trust is given by $\mu _{k}^{a,b} =\alpha _{k}^{a,b} /\left( \alpha _{k}^{a,b} +\beta _{k}^{a,b}\right)$. 
% \begin{equation} \label{eq:trust_expected} 
% \mu _{k}^{a,b} =\alpha _{k}^{a,b} /\left( \alpha _{k}^{a,b} +\beta _{k}^{a,b}\right)
% \text{.}
% \end{equation}
Here we note that $t_{k}^{a,b}$ is the self-reported trust given by the agent $a$, which has some randomness due to subjectivity, while $\mu_{k}^{a,b}$ is the expected trust determined by the experiences.

% Eq.~\eqref{eq:trust_def} captures the characteristics of trust that human trust is affected by interaction experiences and gaining experiences decreases the variance of trust. 
% Note that agent $b$ here can be either a human agent or a robot agent, and hence the definition applies to both human-to-human trust and human-to-robot trust.

\smallskip
\noindent \textbf{Trust update through direct experience.} Similar to~\cite{Guo2020_IJSR}, we update the direct trust experience at time ${k}$ by setting 
\begin{equation}\label{eq:direct_update}
\begin{aligned}
\alpha _{k}^{a,b} & =\alpha _{k-1}^{a,b} +s^{a,b} \cdot p_{k}^{b}\\
\beta _{k}^{a,b} & =\beta _{k-1}^{a,b} +f^{a,b} \cdot \overline{p}_{k}^{b}
% \alpha _{b}^{a} (k) & =\alpha _{b}^{a} (k-1)+s_{b}^{a} \cdot p_{b} (k)\\
% \beta _{b}^{a} (k) & =\beta _{b}^{a} (k-1)+f_{b}^{a} \cdot \overline{p}_{b} (k)
\end{aligned}\text{.}
\end{equation}
Here $p_{k}^{b}$ and $\overline{p}_{k}^{b}$ are the measurements of $b$'s success and failure during time $k$, respectively; $s^{a,b}$ and $f^{a,b}$ are $a$'s unit experience gains with respect to success or failure of $b$, respectively. We require $s^{a,b}$ and $f^{a,b}$ to be positive to ensure that cumulative experiences are non-decreasing. The updated trust $t_{k}^{a,b}$ follows the distribution $\operatorname{Beta} (\alpha _{k}^{a,b} ,\beta _{k}^{a,b})$.

\smallskip
\noindent \textbf{Trust update through indirect experience propagation.} Let $x$ and $y$ denote two human agents and let $A$ denote a robot agent, as illustrated in Fig.~\ref{fig:trust-transfer}. At time $k$, $y$ communicates his or her trust $t_{k}^{y,A}$ on $A$ with $x$, and then $x$ updates his or her indirect trust experience by 
\begin{equation}\label{eq:indirect_update}
\begin{aligned}
\alpha _{k}^{x,A} & =\alpha _{k-1}^{x,A} +\hat{s}^{x,A} \cdot t_{k}^{x,y} \cdot \left[ t_{k}^{y,A} -t_{k-1}^{x,A}\right]^{+}\\
\beta _{k}^{x,A} &  =\beta _{k-1}^{x,A}  +\hat{f}^{x,A} \cdot t_{k}^{x,y} \cdot \left[ t_{k-1}^{x,A} -t_{k}^{y,A}\right]^{+}
\end{aligned}\text{,}
\end{equation}
where the superscript `$+$' means taking the positive part of the corresponding number, i.e., $t^+=\max\{0,t\}$ for a real number $t$, and $t_{k}^{x,A}\sim \operatorname{Beta} ( \alpha _{k}^{x,A} ,\beta _{k}^{x,A} )$. 

The intuition behind this model is that $x$ needs to reason upon $t_{k}^{y,A}$, i.e., $y$'s trust towards $A$. First, $x$ compares $y$'s trust $t_{k}^{y,A}$ with his or her previous trust $t_{k-1}^{x,A}$. Let $\Delta t:=t_{k}^{y,A} -t_{k-1}^{x,A}$ be the difference. If $\Delta t \ge 0$, $x$ gains positive indirect experience about $A$, which amounts to the product of the trust difference $\Delta t$, a coefficient $\hat{s}^{x,A}$, and a discounting factor $t_{k}^{x,y}$, i.e., $x$'s trust on $y$; if $\Delta t<0$, then $x$ gains negative indirect experience about $A$, which is defined similarly. 

% \vspace{-3mm}
\subsection{Parameter Inference}\label{sec:para_infer}
The proposed model characterizes a human agent's trust on a robot by six parameters. For instance, the parameter of $x$ on robot $A$, which is defined as $\theta ^{x,A} =\left( \alpha _{0}^{x,A} ,\beta _{0}^{x,A} ,s^{x,A} ,f^{x,A} ,\hat{s}^{x,A} ,\hat{f}^{x,A}\right)$,
% \begin{equation}
% \label{eq:theta_def}
% \theta ^{x,A} =\left( \alpha _{0}^{x,A} ,\beta _{0}^{x,A} ,s^{x,A} ,f^{x,A} ,\hat{s}^{x,A} ,\hat{f}^{x,A}\right) ,
% \end{equation}
including $x$'s prior experiences $\alpha _{0}^{x,A}$ and $\beta _{0}^{x,A}$, the unit direct experience gains $s_{A}^{x}$ and $f_{A}^{x}$, and the unit indirect experience gains $\hat{s}_{A}^{x}$ and $\hat{f}_{A}^{x}$. We denote the indices of $x$'s direct and indirect trust updates with $A$ up to time $k$ as $D_{k}$ and $\overline{D}_{k}$, respectively. Then, we can compute $\alpha _{k}^{x,A}$ and $\beta _{k}^{x,A}$, according to Eqs.~\eqref{eq:direct_update} and \eqref{eq:indirect_update}, as
\begin{equation}
\label{eq:alpha_k_beta_k}
\begin{aligned}
\alpha _{k}^{x,A} = & \alpha _{0}^{x,A} +s^{x,A}\sum _{j\in D_{k}} p_{j}^{A} +\hat{s}\sum _{j\in \overline{D}_{k}} t_{j}^{x,y}\left[ t_{j}^{y,A} -t_{j-1}^{x,A}\right]^{+}\\
\beta _{k}^{x,A} = & \beta _{0}^{x,A} +f^{x,A}\sum _{j\in D_{k}}\overline{p}_{j}^{A} +\hat{f}\sum _{j\in \overline{D}_{k}} t_{j}^{x,y}\left[ t_{j-1}^{x,A} -t_{j}^{y,A}\right]^{+}
\end{aligned} .
\end{equation}
The optimal parameter $\theta _{*}^{x,A}$ maximizes the log likelihood function 
\begin{equation}
\label{eq:likelihood_def}
H\left( \theta ^{x,A}\right) :=\sum _{k=0}^{K}\log\operatorname{Beta}\left( t_{k}^{x,A}\middle| \alpha _{k}^{x,A} ,\beta _{k}^{x,A}\right) ,
\end{equation}
where $\alpha _{k}^{x,A}$ and $\beta _{k}^{x,A}$ are defined in Eq.~\eqref{eq:alpha_k_beta_k}.
% We compute the optimal parameter $\theta ^{x,A}_{*}$ by maximum likelihood estimation (MLE), i.e., $\theta^{x,A}_{*} =  \arg\max\log\Pr\left(\text{data}\middle| \theta ^{x,A}\right)\\=  \arg\max\sum _{k=0}^{K}\log\operatorname{Beta}\left( t_{k}^{x,A}\middle| \alpha _{k}^{x,A} ,\beta _{k}^{x,A}\right)$.
% \begin{equation*}
% \begin{aligned}
% \theta^{x,A}_{*} = & \arg\max\log\Pr\left(\text{data}\middle| \theta ^{x,A}\right)\\
% = & \arg\max\sum _{k=0}^{K}\log\operatorname{Beta}\left( t_{k}^{x,A}\middle| \alpha _{k}^{x,A} ,\beta _{k}^{x,A}\right)\text{.}
% \end{aligned}
% \end{equation*}
% In particular, we formulate the parameter inference problem as follows: given $x$'s trust history on $A$, $\{t_{k}^{x,A}\}_{k=0,1,\dotsc ,K}$, $A$'s performance history during $x$'s direct interactions, $\{( p_{k}^{A} ,\overline{p}_{k}^{A})\}_{k\in D_{K}}$, $x$'s trust on $y$, $\{t_{k}^{x,y}\}_{k\in \overline{D}_{K}}$, and $y$'s trust on $A$, $\{t_{k}^{y,A}\}_{k\in \overline{D}_{K}}$ during $x$'s indirect trust update, we compute the parameter $\theta _{*}^{x,A}$ that maximizes the log likelihood function 
% \begin{equation}
% \label{eq:likelihood_def}
% H\left( \theta ^{x,A}\right) :=\sum _{k=0}^{K}\log\operatorname{Beta}\left( t_{k}^{x,A}\middle| \alpha _{k}^{x,A} ,\beta _{k}^{x,A}\right) ,
% \end{equation}
% where $\alpha _{k}^{x,A}$ and $\beta _{k}^{x,A}$ are defined in Eq.~\eqref{eq:alpha_k_beta_k}.

We note that $\log\operatorname{Beta}( t_{k}^{x,A} | \alpha _{k}^{x,A} ,\beta _{k}^{x,A})$ is concave in $\theta ^{x,A}$ by the composite rule that the function is concave in $( \alpha _{k}^{x,A} ,\beta _{k}^{x,A})$ and $\alpha _{k}^{x,A}$ and $\beta _{k}^{x,A}$ are non-decreasing linear functions of $\theta ^{x,A}$. Consequently, $H( \theta ^{x,A})$ is concave in $\theta ^{x,A}$ because it is a summation of several concave functions. Therefore, we can run the gradient descent method to compute the optimal parameters.

\section{Human-Subject Study}\label{sec:humanStudy}
We conducted a human-subject experiment to evaluate the proposed model. The experiment, inspired by~\cite{yang2017evaluating}, simulated a threat detection task, where two human agents work with two smart drones to search for threats at multiple sites.   

\subsection{Participants}
A total of $N=30$ participants (average age = 25.3 years, SD = 4.3 ages, 16 females, 14 males) with normal or corrected-to-normal vision formed 15 teams and participated in the experiment. Each participant received a base payment of \$15 and a bonus of up to \$10  depending on their team performance. To promote cooperation between a pair of players, team performance instead of individual performance was used to calculate the bonus. 

\subsection{Experimental Task and Design}
\begin{figure}[!t]
  \centering
    \begin{subfigure}{1\columnwidth}
    \captionsetup{width=1\linewidth}
        \centering
        \includegraphics[width=1\columnwidth]{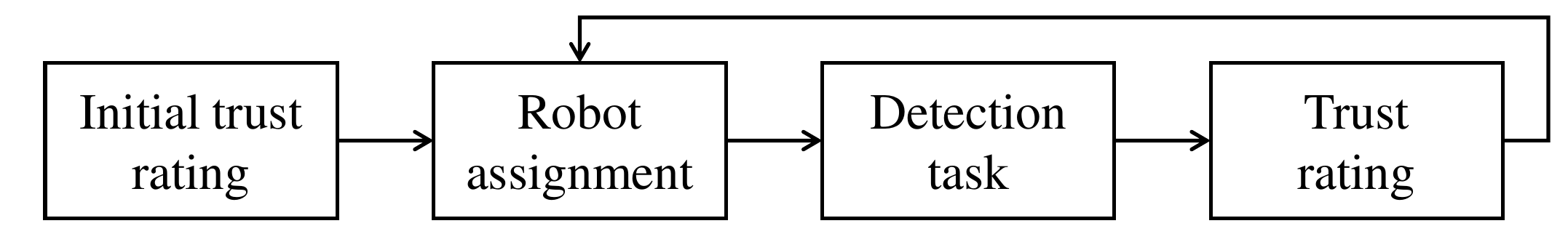}
        \caption{Flow of experimental task}
        \label{fig:process}
    \end{subfigure}
    \begin{subfigure}{1\columnwidth}
    \captionsetup{width=1\linewidth}
        \centering
        \vspace{5pt}
        \includegraphics[width=1\columnwidth]{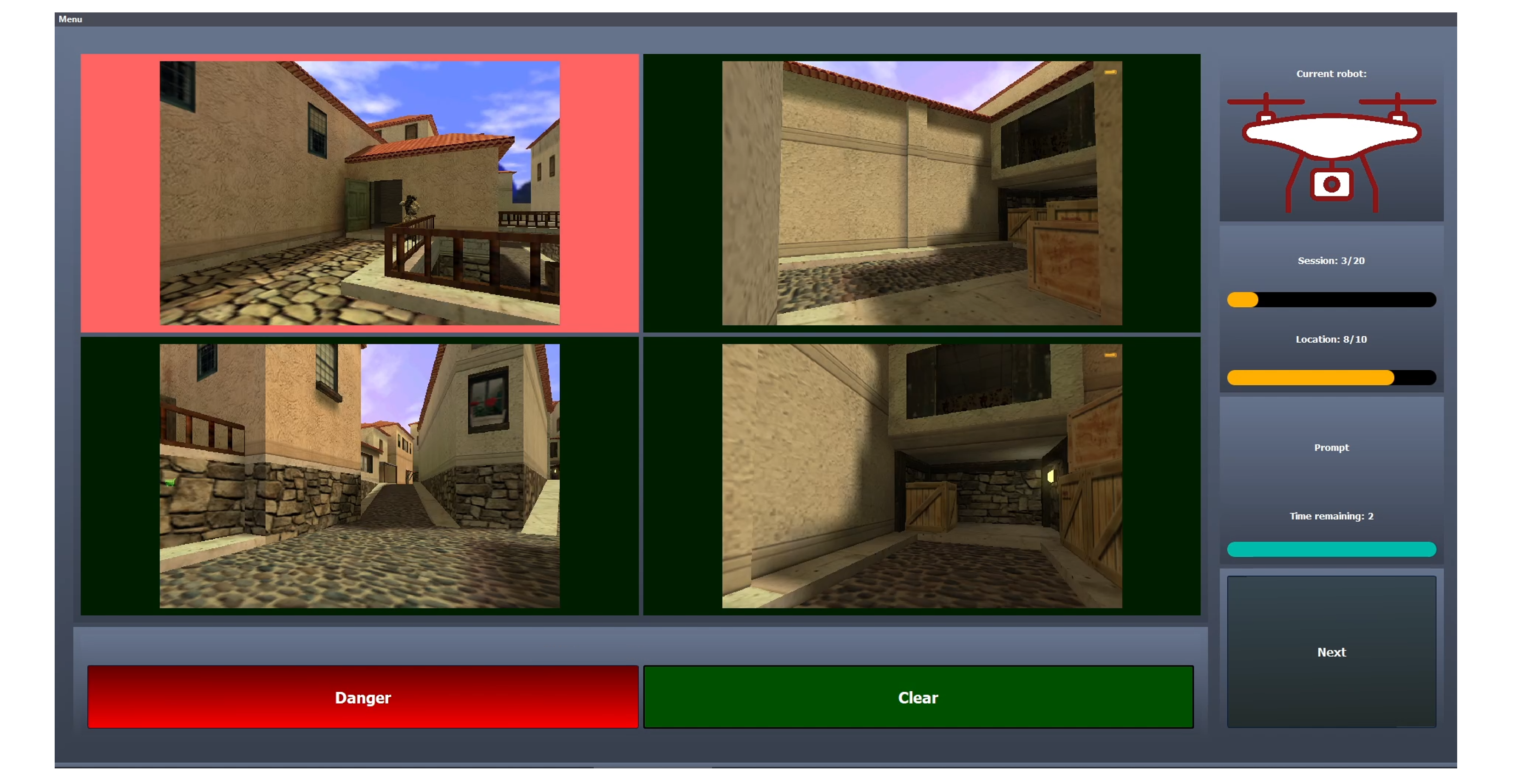}
        \caption{Task interface. The drone will highlight the potential threat in bright red. The participant is asked to click the `Danger' button if a threat is present and to click the `Clear' button otherwise.}
        \label{fig:interface}
    \end{subfigure}
    
    % \begin{subfigure}{1\columnwidth}
    % \captionsetup{width=1\linewidth}
    %     \centering
    %     \vspace{10pt}
    %     \includegraphics[width=1\columnwidth]{figs/trust-rating.pdf}
    %     \caption{After each session, each participant updates his or her direct trust on the drone s/he just worked with (80 and 60 in the figure). And then the system shows the direct trust values to their human teammates and asks them to update their indirect trust (70 and 75 in the figure). Finally, the participants update their trust towards each other (90 and 85 in the figure).}
    %     \label{fig:trust-rating}
    % \end{subfigure}
    
    \caption{Experimental task and design}
  \label{fig:experiment_design}
\end{figure}

In the experiment, a pair of participants performed a simulated threat detection task with two assistant drones for $K=15$ sessions on two separate desktop computers. At each session, each participant was assigned one drone and worked on the detection tasks. After the session, they were asked to report their trust in each drone and their trust in their human teammate. 
 % participant's trust towards each drone,  the second type was each participant's trust towards the other participant, which was based on how much they would trust on the other participant's ability to rate trust on the drones. 
 For clarity, we named the two drones $A$ and $B$ and colored them in red and blue, respectively; and we denoted the participants $x$ and $y$. A trust rating is denoted as $t^{a,b}_k$, where the superscript $a\in \{x,y\}$ stands for the trustor, the superscript $b\in \{x,y,A,B\}$ stands for the trustee, and the subscript $k$ is the session index. For example, $t^{x,A}_2$ is person $x$'s trust in drone $A$ after the 2nd session. The range of a trust rating is $[0,1]$, where 0 stands for ``(do) not trust at all'' and 1 stands for ``trust completely''. The flow of the experimental task is illustrated in Fig.~\ref{fig:process}.

\smallskip
\textbf{Initial trust rating}: At the start, each participant gave their initial trust in the two drones based on their prior experience with automation/robots. Additionally, they gave their initial trust in each other. These trust ratings were indexed by 0, e.g., $x$'s initial trust rating on $A$ was denoted as $t^{x,A}_0$.

\smallskip
\textbf{Robot assignment}: At each session, each participant was randomly assigned one drone as his or her assistant robot.

\smallskip
\textbf{Detection task}: Each session consisted of 10 locations to detect. As shown in Fig.~\ref{fig:interface}, four views were present at each location. If a threat, which appeared like a combatant, was in any of the views, the participant should click the `Danger' button; otherwise, they should click the `Clear' button. Meanwhile, his or her drone would assist and highlight a view if the drone detected a threat there. In addition, a 3-second timer was set for each location. If a participant did not click either button before the timer counted down to zero, the testbed
% would mark his or her detection result as wrong for the current location and then 
would move to the next location automatically. After all the 10 locations, an end-of-session screen was shown, displaying how many correct choices the participant and the drone had made in the current session. Correct choices mean correctly identifying threats or declaring `Clear' within 3 seconds. 
% Here the rule was that, at each location, a participant's choice would be marked as wrong if he or she clicked the wrong button or did not click either button before time was out; otherwise, it was marked as correct. For the drone, its choice was wrong if its detection result was wrong, e.g., it highlighted a view without a threat.
\begin{figure*}[t]
  \centering
  \includegraphics[width=1\linewidth]{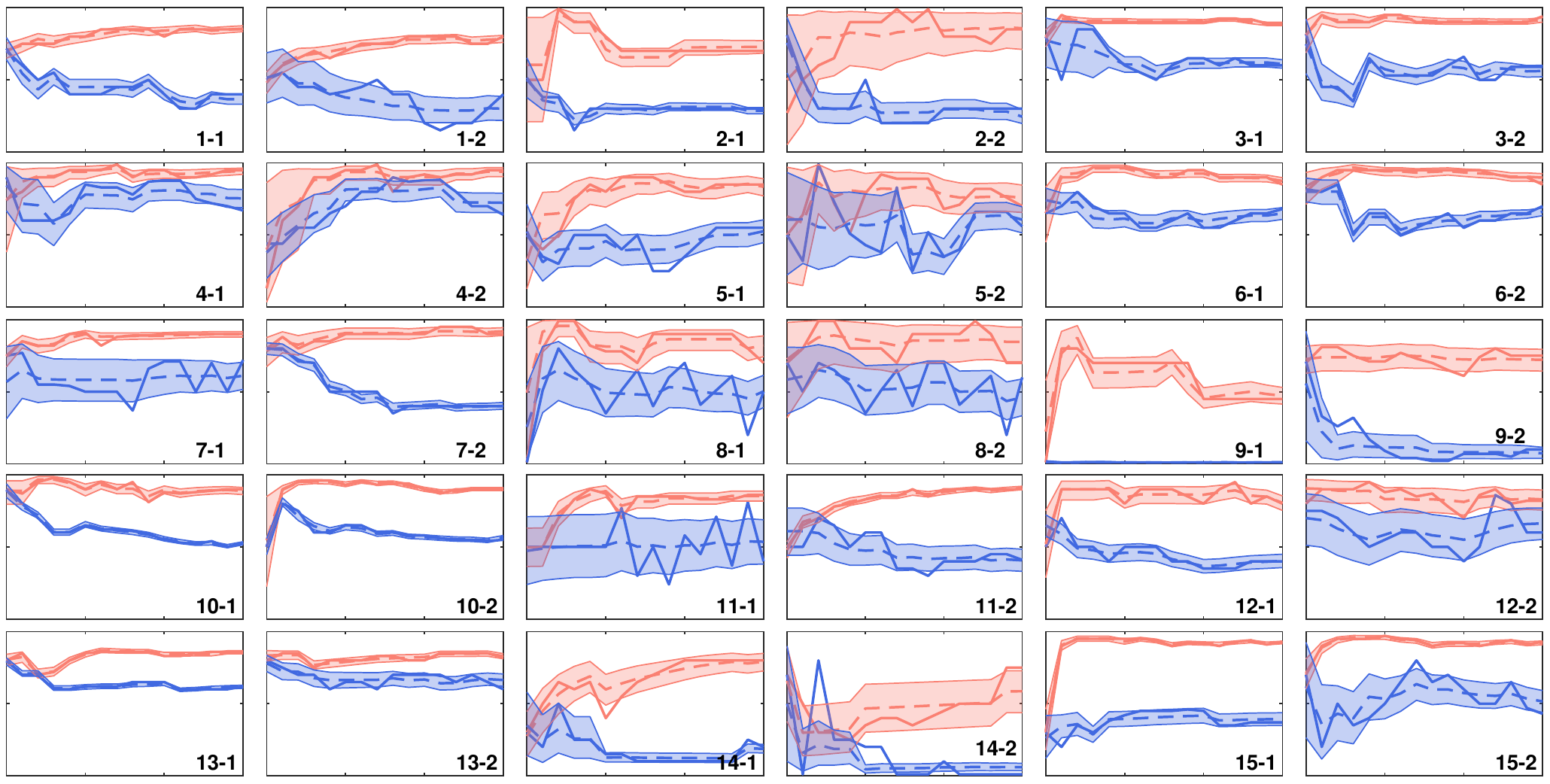}
    \vspace{-2mm}
  \caption{Fitting results. Red curves are for drone $A$ while blue curves are for drone $B$. The solid lines are the participants' self-reported trust, while the dashed lines are the expected trust value predicted by the model. The shaded areas indicate the 90\% probability interval of the Beta distribution at each session. The index $i$-$j$ stands for the $j$th participant in the $i$the group.}
  \vspace{-2mm}
  \label{fig:fitting_all}
\end{figure*}

\textbf{Trust rating}: After each session, participants reported three trust values.
% , as shown in Fig.~\ref{fig:trust-rating}. 
First, each participant updated his or her trust in the drone s/he just worked with, i.e., through direct experience. Next, each participant submitted and communicated their trust score to their human teammate. After that, each participant updated his or her trust in the drone the human teammate just worked with (i.e., the other drone) and his or her trust in the human teammate. After participants completed all 15 sessions, the experiment ended.

% Next, the system showed these direct trust ratings to both the participants and asked for their \textit{indirect trust update}, e.g., each participant's trust rating on the drone that he or she did not work with in the current session. Finally, the participants updated their trust towards each other. 

 % otherwise, they went to the robot-assignment step to start the next session.

% After completing all 15 sessions, the participants received a bonus determined by their average detection accuracy, designed to encourage them to give their true trust feedback. The participants were told that giving true trust feedback would help his or her human teammate to decide how much s/he should trust the drones and thus increase their detection accuracy. 
% ============ the big spanning figure ================

\subsection{Experimental Procedure}
Before the experiment, each participant signed a consent form and filled out a demographic survey. Two practice sessions were provided, wherein a practice drone was used to assist the participants. The participants were told that the practice drone differed from the two drones used in the real experiment. After the experiment started, the assignment of drones was randomized in each group. Specifically, we assigned drone $A$ with equal change to either participant and then assigned drone $B$ to the other participant. The threat detection accuracy of the practice drone, drone $A$, and drone $B$ were set to 80\%, 90\%, and 60\%, respectively.

\section{Results and Discussion}
We use the gradient descent method in Sec.~\ref{sec:para_infer} to compute the optimal parameters $\theta^{p_i,A}_*$ and $\theta^{p_i,B}_*$ for each participant $p_i$. The fitting results are shown in Fig.~\ref{fig:fitting_all}. We calculate the performance measurements of drone $A$ at session $k$ as $p_{k}^{A} =A_{k} /10$ and $\overline{p}_{k}^{A} =1-p_{k}^{A}$, where $A_{k}$ is the number of correct choices drone $A$ made in the $k$th session; and we define $p_{k}^{B}$ and $\overline{p}_{k}^{B}$ similarly. To measure the performance of the model, we calculate the fitting error at each session for each participant as 
$e_{k}^{p_{i} ,R} =| \mu _{k}^{p_{i} ,R} -t_{k}^{p_{i} ,R}|$, $R\in \{A,B\}$, 
% \begin{equation*}
% % e_{k}^{p_{i} ,R} =\left| \mu _{k}^{p_{i} ,R} -t_{k}^{p_{i} ,R}\right| ,\ R\in \{A,B\},
% {\textstyle e_{k}^{p_{i} ,R} =\left| \mu _{k}^{p_{i} ,R} -t_{k}^{p_{i} ,R}\right| ,\ R\in \{A,B\},}
% \end{equation*}
where $t_{k}^{p_{i} ,R}$ is the participant's self-reported trust while $\mu_{k}^{p_{i} ,R}$ is the expected trust defined in section~\ref{sec:model} and computed based on $\theta _{*}^{p_{i} ,R}$;
% computed according to Eq.~\eqref{eq:trust_expected} with $\alpha _{k}^{p_{i} ,R}$ and $\beta _{k}^{p_{i} ,R}$ generated by Eq.~\eqref{eq:alpha_k_beta_k} based on $\theta _{*}^{p_{i} ,R}$; 
and, we calculate the root-mean-square error (RMSE) between the ground truth and the expected trust value as
\begin{equation*}
% \text{RMSE}^{R} =\left[\frac{1}{N}\sum _{i=1}^{N}\frac{1}{K+1}\sum _{k=0}^{K}\left( e_{k}^{p_{i} ,R}\right)^{2}\right]^{1/2} ,
{\textstyle \text{RMSE}^{R} =\left[\frac{1}{N}\sum _{i=1}^{N}\frac{1}{K+1}\sum _{k=0}^{K}\left( e_{k}^{p_{i} ,R}\right)^{2}\right]^{1/2} ,}
\end{equation*}
%where $N=30$ is the number of participants, $K=15$ is the number of sessions for each participant, and 
for $R\in \{A,B\}$. The RMSE results for the TIP model are $\text{RMSE}^A=0.057$ and $\text{RMSE}^B=0.082$.

Fig.~\ref{fig:fitting_all} shows the fitting results of the TIP model. The shaded regions indicate the 90\% confidence interval of the Beta distribution at each session. We observe that for most participants, such as 7-2 and 10-2, the proposed TIP model can accurately fit the trust curve with a narrow confidence interval; but for some other participants, such as 5-2 and 8-1, the model cannot fit the trust curve due to trust oscillation. However, in the latter case, the fitted curve has a similar trend with the ground truth and can cover most data points with the 90\% confidence interval. 

For comparison, we consider a direct-update-only model that only accounts for the direct experience a human agent has with a robot. The direct-update-only model is equivalent to the TIP model with zero unit indirect experience gains, i.e., $\hat{s}^{x,A}=\hat{f}^{x,A}=0$. We recompute the model parameters for the direct-update-only model, and the corresponding RMSE errors are $\text{RMSE}'^A=0.085$ and $\text{RMSE}'^B=0.107$. Furthermore, we compare each participant's fitting error $\bar{e}^{p_{i} ,R}:={1}/({K+1})\sum_{k=0}^K e_{k}^{p_{i} ,R}$ of the TIP model ($A$: $0.044 \pm 0.037$; $B$: $0.069 \pm 0.045$) and that of the direct-update-only model ($A$: $0.075 \pm 0.041$; $B$: $0.095 \pm 0.051$) using a paired-sample t-test. Results show that the former is significantly smaller than the latter, with $t(29)=-6.18, p<.001$ for drone $A$, and $t(29)=-7.31$, $p<.001$ for drone $B$.  

% Furthermore, we compare each participant's fitting error $\bar{e}^{p_{i} ,R}:={1}/({K+1})\sum_{k=0}^K e_{k}^{p_{i} ,R}$ of the TIP model ( $0.0435 \pm 0.0368$ on $A$ and $0.0687 \pm 0.0453$ on $B$) and that of the direct-update-only model ($0.0752 \pm 0.0414$ on $A$ and $0.0954 \pm 0.0505$ on $B$) using a paired-sample t-test. Results show that the former is significantly smaller than the latter, with $t(29)=-6.1810, p<0.001$ on $A$ and $t(29)=-7.3147$, $p<0.001$ on $B$.  

% Further comparison shows that, for both drone $A$ and drone $B$, each participant's fitting errors $e^{p_i,R}_k$ of the TIP model are significantly smaller than the ones of the direct-update-only model ($p<0.001$ for both drones, t-test). NEED TO CHECK WITH YAOHUI!!!
% (paired t-test, $t(29)=XXXX$, $p<XXX$)
% \vspace{-5mm}
\section{Acknowledgement}
This work is supported by the National Science Foundation under Grant No. 2045009 and the Air Force Office of Scientific Research under Grant No. FA9550-20-1-0406.

\bibliographystyle{ACM-Reference-Format}
\bibliography{HFES-bibliography}


% \newpage
% \appendix

% \section{Participants $x$'s and $y$'s trust in drones $A$ and $B$ by session}

% Participant $x_i$' and $y_i$'s trust in drone $A$ and drone $B$. $i = 1, 2, \dots, 15$ denotes the team number.  The red line indicates drone $A$ and the blue line drone $B$. The same color between a shaded area and a line indicates \textit{direct} experience, otherwise \textit{indirect} experience. 

% \begin{figure}[H]
%     \centering
%     % \includegraphics[width=0.7\columnwidth]{images/fig_trust_with_direct_background.pdf}
%     % \includegraphics[width=1\columnwidth]{images/bigFigure_8by4.pdf}
%     \includegraphics[width=1\columnwidth]{images/bigFigure_8by4.pdf}
%     % \caption{Trust plots of all 30 participants in the experiments.}
% \label{fig:all_plots2}
% \end{figure}

\end{document}